\documentclass[12pt]{article}
\usepackage[top=1in, bottom=1in, left=1.0in, right=1.0in]{geometry} 
\usepackage{color}
\usepackage{enumitem}
\usepackage{cite}
\usepackage{hyperref}
\hypersetup{
	colorlinks = true,
	allcolors = blue,
}

\definecolor{kentuckyblue}{RGB}{0, 93, 170}			
\definecolor{green}{RGB}{0, 102, 0}					
\definecolor{frenchred}{RGB}{250,60,50}				
\definecolor{orange}{RGB}{255,127,0}             

\definecolor{purple}{RGB}{204, 0, 255}					

\title{Ethical Considerations in Artificial Intelligence Courses}
\author{
Emanuelle Burton \\ 
University of Kentucky \\
\textit{emanuelle.burton@gmail.com}
\and Judy Goldsmith \\
University of Kentucky \\
\textit{goldsmit@cs.uky.edu}
\and Sven Koenig \\ 
University of Southern California\\
\textit{skoenig@usc.edu}
\and Benjamin Kuipers \\
University of Michigan\\
\textit{kuipers@umich.edu }
\and Nicholas Mattei \\
IBM T.J. Watson Research Center \\
\textit{n.mattei@ibm.com}
\and Toby Walsh \\
UNSW, Data61/CSIRO, and TU Berlin\\
\textit{toby.walsh@data61.csiro.au}
}
\date{}

\begin{document}

\maketitle


\begin{abstract}
The recent surge in interest in ethics in artificial intelligence may leave many educators wondering how to address moral, ethical, and philosophical issues in their AI courses. As instructors we want to develop curriculum that not only prepares students to be artificial intelligence practitioners, but also to understand the moral, ethical, and philosophical impacts that artificial intelligence will have on society. In this article we provide practical case studies and links to resources for use by AI educators.  We also provide concrete suggestions on how to integrate AI ethics into a general artificial intelligence course and how to teach a stand-alone artificial intelligence ethics course.
\end{abstract}

\section{Introduction}

Artificial Intelligence is one of the most ambitious scientific and engineering adventures of all time.  The ultimate goal is to understand the mind from a new perspective, and to create AIs\footnote{We create artifacts which take multiple forms including intelligent computing systems and robots.  In this article we use the term AI and AIs to refer to any artificial, autonomous decision maker.} capable of learning and applying intelligence to a wide variety of tasks: some as robots able to take action in our physical and social world, and some as software agents that make decisions in fractions of a second, controlling huge swaths of the economy and our daily lives.  However, the power and reach of these AIs makes it necessary that we consider the risks as well as the rewards. 

In thinking through the future of AI, it is useful to consider fiction, especially science fiction. From Frankenstein's monster and Hoffmann's automata to Skynet and Ex Machina, fiction writers have raised concerns about destruction that could perhaps be unleashed on humanity by the autonomy we confer on our technological creations. What is the underlying concern that inspires so many variations of this story? Do these storytellers (and their audiences) fear that AIs, by definition, cannot be trusted to act in a way that does not harm the society that creates it? Or do they fear that the people in charge of designing them are making the wrong choices about how to design them in the first place? 

For these storytellers and their audiences, AIs may be a narrative device (so to speak) for thinking about basic questions of ethics; but they can also help AI designers and programmers to think about the risks, possibilities and responsibilities of designing autonomous decision makers.  As Peter Han argues in his dissertation \cite{han2015towards}, we cannot simply slap on an ethics module after the fact;  we must build our systems from the ground up to be ethical.  But in order to do that, we must also teach our AI programmers, practitioners, and theorists to consider the ethical implications of their work.

Recent dramatic progress in AI includes programs able to achieve super-human performance at difficult games like Jeopardy and Go; self-driving cars able to drive on highways and city streets with an excellent (though certainly not flawless) safety record; and software that enables face, speech, and activity recognition across unimaginably large data sets \cite{FaceRec}. These advances have prompted various public figures and thinkers to raise questions about the possible threats that AI research and applications could pose to the future of humanity.  Even without a looming apocalypse, however, the concerns are real and pressing.  When intelligent systems interact with humans they are functioning, at least in part, as members of society.  This integration of AI into our lives raises a number of interesting and important questions, large and small, of which we give a brief overview in the next section.  These questions -- as well as any answers we might supply to them -- are ethical as well as practical, and the reasoning structures that we use to identify or answer them have deep affinities to the major traditions of ethical inquiry.

Engineering education often includes units, and even entire courses, on professional ethics, which is the ethics that human practitioners should follow when acting within their profession.  Advances in AI have made it necessary to expand the scope of how we think about ethics; the basic questions of ethics -- which have, in the past, been asked only about humans and human behaviors -- will need to be asked about human-designed artifacts, because these artifacts are (or will soon be) capable of making their own action decisions based on their own perceptions of the complex world. How should self-driving cars be programmed to choose a course of action, in situations in which harm is likely either to passengers or to others outside the car?  This kind of conundrum -- in which there is no ``right'' solution, and different kinds of harm need to be weighed against each other -- raises practical questions of how to program a system to engage in ethical reasoning. It also raises fundamental questions about what kinds of values to assign to help a particular machine best accomplish its particular purpose, and the costs and benefits that come with choosing one set of values over another; see, for instance, \cite{bonnefon-arxiv,MITTR:bonnefon} for discussion of this issue.

Just as students of AI learn about search, knowledge representation, inference, planning, learning, and other topics, they should also learn about ethical theories: how those theories help name, explain, and evaluate the inclinations and principles that already inform our choices, and how those theories can be used to guide the design of intelligent systems.  We here provide a brief primer on basic ethical perspectives, and offer a series of case studies that show how these viewpoints can be used to frame the discourse of how AIs are designed, built, and understood.  These case studies can be used as a jumping off point for including an ethics unit within a university course of Artificial Intelligence.

\section{Some Ethical Problems Raised by AIs}

The prospect of our society including a major role for AIs poses numerous deep, important questions, many of which can be understood and analyzed through the lens of ethical theory.  We briefly discuss a few of these issues which have recently received the most attention, recognizing that there are many others \cite{Russell-aimag-15}; we will come back to a number of these questions by integrating them with specific case studies for use in the classroom.

\subsection{How Should AIs Behave in Our Society?}
AIs at their most basic level are computer programs that are capable of making decisions.  While currently these systems are mostly software agents responsible for approving home loans or deciding to buy or trade stocks, in the future these AIs could be embodied, thus perceiving and acting in the physical world.  We all know that computer programs can have unintended consequences and embodied computer systems raise additional concerns.  Fiction raises apocalyptic examples like SkyNet in the Terminator movies, but real-world counterparts such as high-speed algorithmic trading systems have actually caused ``flash crashes" in the real economy \cite{kirilenko2015flash}. We address topics related to both of these in the SkyNet and Machine Learning case studies.

We can also expect robots to become increasingly involved in our daily lives, whether they are vacuuming our floors, driving our cars, or helping to care for our loved ones.  How do their responsibilities for these tasks relate to other ethical responsibilities to society in general?  We address this in the Robot \& Frank case study below.

\subsection{What Should We Do if Jobs Are in Short Supply?}
As AIs become more powerful, the traditional economy may decrease the number of jobs for which human workers are competitive, which could increase inequality, thus decreasing the quality of our economy and our lives.  Alternatively, our society could recognize that there are plenty of resources, plenty of work we want done, and plenty of people who want to work.  We could take an approach that deliberately allocates resources to provide jobs that are not currently justified by increasing shareholder profits, but will improve the quality of life in our society.  This topic already receives a great deal of attention from computer scientists, economists, and political scientists (e.g.,  \cite{EconomistImpact,Piketty-14,brynjolfsson2014second,ford2015rise,Schumacher-79}).  Therefore, although we certainly grant its importance, we do not pursue this topic further in this paper.

\subsection{Should AI Systems Be Allowed to Kill?}
There are several ethical arguments, and a popular movement, against the use of killer robots in war (see, e.g., \url{http://futureoflife.org/open-letter-autonomous-weapons/}).  Critics of killer robots argue that developing killer robots will inevitably spark a global arms race, and that there will be no way to prevent repressive governments, terrorist groups or ethnic cleansing movements from acquiring and using this technology once it exists.  They argue, further, that there are ways to use AI in warfare that are not about killing. There are also a number of arguments in favor of robots that kill. Advocates of robots that kill claim that some wars are necessary and just; that killer robots will take humans out of the line of fire; that such robots can be used for deterrence as well as for actual violence; and that it is unrealistic to try to prevent this technology, since it already exists in some forms, and there are significant political and financial resources devoted to making sure that it be developed further.  It is further argued that robots will be better than humans at following the laws of war and the rules of engagement that are intended to prevent war crimes \cite{Arkin-09}.  The question of robots that kill has been receiving a lot of attention from various institutes including the Future of Life Institute (\url{http://futureoflife.org/}) and The Campaign to Stop Killer Robots (\url{https://www.stopkillerrobots.org/}).  We address this case tangentially with the Skynet case study but we do not engage directly with the morality of war.

\subsection{Should We Worry About Superintelligence and the Singularity?}
Following the highly influential book \emph{Superintelligence} by Nick Bostrom \cite{Bostrom-14}, several high profile scientists and engineers expressed concerns about a future in which AI plays a key part. Elon Musk called artificial intelligence our ``biggest existential threat.''\footnote{\url{http://webcast.amps.ms.mit.edu/fall2014/AeroAstro/index-Fri-PM.html}}  Bill Gates was a little more circumspect, stating ``I am in the camp that is concerned about super intelligence. First, the machines will do a lot of jobs for us and not be super intelligent. That should be positive if we manage it well. A few decades after that, though, the intelligence is strong enough to be a concern.''\footnote{\url{http://www.businessinsider.com/bill-gates-artificial-intelligence-2015-1}} Tom Dietterich gave a presentation at the 2015 DARPA Future Technology workshop in which he argued against fully autonomous AI systems.\footnote{\url{https://www.youtube.com/watch?v=dQOo3Mg4D5A}}  Vince Conitzer also discussed the reasons that many AI practitioners do not worry about the Singularity \cite{conitzer:prospect}; The Singularity is likely to be a low-probability problem, compared with the others discussed in this section, but obviously the stakes are extremely high.  There is a wealth of resources detailing the ethical considerations we have to the machines, and ourselves, from a number of concerned institutions including the Future of Humanity Institute (\url{https://www.fhi.ox.ac.uk/}) and the Machine Intelligence Institute (\url{https://intelligence.org/}), among others mentioned already.

\subsection{How Should We Treat AIs?}
As AIs are embedded more fully into our society, we will face again a pressing ethical dilemma that has arisen repeatedly throughout the centuries: how do we treat ``others''?   Some of the groups that have been classed as ``others'' in the past include animals (endangered species in particular), children, plants, the mentally disabled, the physically disabled, societies that have been deemed ``primitive" or ``backward," citizens of countries with whom we are at war, and even artifacts of the ancient world.  Currently, the EPSRC Principles of Robotics \cite{EPSRC-Principles-11}, along with other leading scholars including Joanna Bryson \cite{bryson2010robots}, are very clear on this topic:  robots are not the sort of thing that have moral standing.  While the current state of technology makes this distinction fairly easy, it is not difficult to imagine a near-term future where robots are able to develop a unique body of knowledge and relationship to that knowledge, and hence may or may not be entitled to more consideration.  This question is touched on by our Robot \& Frank case study.

\section{Tools for Thinking about Ethics and AI}

Ethics as a discipline explores how the world should be understood, and how people ought to act.  There are many schools of thought within the study of ethics, which differ not only in the answers that they offer, but in the ways that they formulate basic questions of how to understand the world, and to respond to the ethical challenges it presents. Most (though not all) work in ethics --- both academically and in the wider world --- has a normative purpose: that is, it argues how people ought to act.  But this normative work relies significantly, though often invisibly, on \emph{descriptive} arguments; before offering prescriptions for how to address a given problem, scholars in ethics construct arguments for why it is both accurate and useful to understand that problem in a particular way. We contend that this descriptive dimension of ethics is as important as the normative, and that instructors should push their students to develop the ability to describe situations in ethical terms, as well as to render judgment.

Most approaches to understanding the world through ethics adopt one of the three major critical orientations: deontological ethics, utilitarianism (sometimes called consequentialism), and virtue ethics. In order to understand and discuss the ethical issues around AIs, it is necessary to be familiar with, at a minimum, these three main approaches. We offer a brief summary of each of these theories here. For a more in-depth examination, there are a number of good resource texts in ethics (e.g., \cite{copp2005oxford,Blackwell}, the Internet Encyclopedia of Philosophy, and the Stanford Encyclopedia of Philosophy) that offer more in-depth and insightful introductions to these and other theories that one could teach in a larger ethics course.  A good discussion of issues in computer science analyzed through ethical theories can be found in \emph{Computer Ethics} \cite{johnson2009computer}.

It is worth noting, up front, that these three approaches need not be, and indeed should not be, treated as independent or exclusive of the others. We are not arguing for the superiority of any particular system; indeed, we believe that a thorough ethics education will equip students to make use of all three major theories, and in some cases to use them in combination. Part of the goal of an AI ethics class should be to teach students to consider each problem from multiple angles, and to reach a considered judgment about which theory (or which theories in combination) are best suited to address a particular problem, and consider the effects of possible solutions.

\subsection{Deontology}
Deontology understands ethics to be about following the moral law. In its most widely-recognized form, it was developed by Immanuel Kant in the late eighteenth century, but has ancient roots in both Divine Command traditions (such as ancient Israelite religion, the source of the Ten Commandments and the basis of Judaism, Christianity and Islam) and in other legal codes. The basic question of deontology is ``what is my duty?''  According to deontology, that duty can be understood in the form of laws. According to Kant, it is the responsibility of every individual to discover the true moral law for him or herself. Although the theoretical rationales for law-based ethics and Kantian deontology are different, in both systems any true law will be universally applicable.
Deontology meshes very well with both specialist and popular understandings of how an ethical machine might come into being.  Isaac Asimov's {\em I, Robot} \cite{asimov1950robot} looks at the consequences of building ethics based on his Three Laws of Robotics.\footnote{An anonymous reviewer suggested that we can summarize Asimov's three laws as decreasing priorities of human-preservation, human-obedience, and robot-self-preservation; the 0th law would be humanity-preservation.}
  Students may perceive deontological analysis to be analogous to application of axiomatic systems.  The underlying questions become, ``How are rules applied to decisions?" and ``What are the right rules?"  The latter question is one of mechanism design, namely, what rules do we put in place in order to achieve our desired social goals? The latter formulation risks departing from deontology, however, unless the desired social goals are brought into alignment with a universal form of justice.

\subsection{Utilitarianism}
The most recent approach, utilitarian ethics, was developed by Jeremy Bentham and John Stuart Mill in the late 18th to mid-19th century. The basic question of utilitarianism is ``what is the greatest possible good for the greatest number?" --- or, in William K. Frankena's more recent formulation \cite{frankena}, ``the greatest possible balance of good over evil." In computer science, and broadly in the social sciences we use ``utility" as a proxy for individual goodness and the sum of individual utilities as a measure of social welfare, often without reflecting on the possibility of thinking about social good in other ways.  The underlying assumption is that utility can be quantified as some mixture of happiness or other qualities, so that we can compare the utilities of individuals, or the utility that one person derives in each of several possible outcomes.  The so-called ``utilitarian calculus'' compares the sum of individual utility (positive or negative) over all people in society as a result of each ethical choice.  While classic utilitarianism does not associate probabilities on possible outcomes, and is thus different from decision-theoretic planning, the notion of calculating expected utility as a result of actions fits well into the utilitarian framework.  Utilitarianism is the foundation for the game-theoretic notion of rationality as selecting actions that maximize expected utility, where utility is a representation of the individual agent's preference over states of the world. As with defining ``everyone'' in consequentialism, defining ``utility'' is the crux of applying game-theoretic rationality, and is a source of many difficulties.

Utilitarianism's influence is felt within many areas of computer science, economics, and decision making broadly construed, through the prevalence of game theory \cite{MSZ13a}.  Game theory is an analytical perspective of mathematics that is often used in AI to understand how individuals or groups of agents will interact.  At the most fundamental level, a game theoretic analysis is consequentialist in nature; every agent is a rational, utility maximizer.  While utility is often used to represent individual reward, it can be used to represent much more sophisticated preferences among states of affairs.  This analytic lens has provided numerous insights and advantages to algorithms that are commonly used on the web and in everyday life. 

\subsection{Virtue Ethics}
Virtue ethics (also known as teleological ethics) is focused on ends or goals. The basic question of virtue ethics is ``who should I be?" Grounded in Aristotle and outlined most clearly in the Nichomachean Ethics, \cite{AristotleNichEth}, virtue ethics is organized around developing habits and dispositions that help a person achieve his or her goals, and, by extension, to help them flourish as an individual \cite{annas}. In contrast to deontological ethics, virtue ethics considers goodness in local rather than universal terms (what is the best form/version of this particular thing, in these particular circumstances?) and emphasizes not universal laws, but local norms.  A central component of living well, according to virtue ethics, is ``phronesis," (often translated as ``moral prudence" or ``practical wisdom"). In contrast to pure knowledge (``sophia''), phronesis is the ability to evaluate a given situation and respond fittingly, and is developed through both education and experience.

Virtue ethics was, for many centuries, the dominant mode of ethical reasoning in the west among scholars and the educated classes. It was eclipsed by utilitarian ethics in the late 18th and 19th centuries, but has seen a resurgence, in the past fifty years, among philosophers, theologians, and some literary critics.  For two thinkers who advance this widely-acknowledged narrative, see \cite{Anscombe2005} and \cite{MacIntyre2007}.

\subsection{Ethical Theory in the Classroom: Making the Most of Multiple Perspectives}
The goal of teaching ethical theory is to better equip our students to understand ethical problems by exposing them to multiple modes of thinking and reasoning. This is best accomplished by helping them understand the powers and limits of each approach, rather than trying to demonstrate the superiority of one approach over the other.  While all three schools have proponents among philosophers, theologians, and other scholars who work in ethics, broader cultural discourse about ethics tends to adopt a utilitarian approach, often without any awareness that there are other ways to frame ethical inquiry.  To paraphrase Ripstein \cite{ripstein:review}, most (American) students, without prior exposure to ethical inquiry, will be utilitarians by default; utilitarianism held unquestioned dominance over ethical discourse in the US and Europe from the mid-19th century until the late 20th, and utilitarianism's tendency to equate well-being with wealth production and individual choice lines up comfortably with many common definitions of `American' values. Studies in other countries, including Italy, show that many students are highly utilitarian in their world views \cite{PCZCS14a}.

This larger cultural reliance on utilitarianism may help explain why it consistently seems, to the students, to be the most crisply-defined and ``usable" of the ethical theories. But there are significant critical shortcomings to utilitarianism, most particularly its in-substantive definition of ``goodness" and the fact that it permits (and even invites) the consideration of particular problems in isolation from larger systems. These shortcomings limit our ability to have substantive ethical discussions, even insofar as everyone assents to utilitarianism; a shared reliance on the principle of ``the greatest good for the greatest number" does not help us agree about what goodness is, or even to reach an agreement about how to define or measure it.  

These same limitations surface in student conversations about ethics. A common problem in their application of utilitarianism is that they may look too narrowly at who is affected by a given decision or action; for example, when considering whether to replace factory workers with robots.  Those making decisions may focus on the happiness of the factory owners, shareholders, and those who can purchase the manufactured goods more cheaply, without considering the utility of the factory workers and those whose jobs depend on factory workers having money to spend; still less are they likely to consider the shortcomings of an ethical model that makes it possible to conceive of human beings and machines as interchangeable. 

A solid education in ethics will teach students about all three approaches to ethics.  This education will allow students to consider a particular problem from a range of perspectives by consider the problem in light of different theories; often the best solution involves drawing on a combination of theories.  For example, in imagining a robot that takes part in human society, students may find it useful to draw upon a combination of deontology and virtue ethics to determine how it is best for that robot to behave, using deontology to establish baseline rules for living, but virtue ethics to consider how the robot could and should incorporate the things it learns.

And yet it is essential that each of these three approaches be taught as distinct from the others. Deontology, utilitarianism and virtue ethics do not represent different ordering systems for identical sets of data; rather, each system offers a profoundly different outlook on meaning and value. It is often the case that the most urgent question, according to one theory, appears by the lights of another theory to be unimportant, or based on flawed premises that are themselves the real problem. 

Consider the question of whether targeted advertising is ethical:  whether it is ethical for advertisers or their service providers to use information harvested from individuals' GPS, email, audio stream, browser history, click history, purchase history, etc., to reason about what goods and services they might be tempted to spend money on.    
The utilitarian analysis takes into account the need for revenue for the provider of free or inexpensive servers and content, plus the utility the user might derive from discovering new or proximally available opportunities, and weighs that against the user's discomfort in having their data shared. Depending on the weight placed on keeping services available to all, and on the business model that requires profit, as well as the utility that individuals are perceived to derive from being exposed to the ads that are selected specifically for them, one might conclude that advertising is a necessary evil, or even a positive. 

The deontological analysis of targeted advertising might focus on the user agreements that allow advertisers access to both the data and the screen real estate, and conclude that legal collection of that data is ethically permissible, given the user agreements.  A virtue ethics analysis might hold as an ideal the ability to work in a focused state, ignoring the visual disturbances. Depending on one's ideal state as a consumer, a virtue ethics model could have the user ignoring the clickbait and advertisements as unworthy, or in following links and even occasionally spending money, so as to support the web of commerce.

The strength of teaching all three systems is that it will equip students to consider the basic nature of ethical problems in a variety of ways. This breadth of perspective will help them confront difficult choices in their work.

Furthermore, students should be discouraged from assuming that the ``best" solution to any given problem is one that lies at the intersection of the three theories. Insightful new solutions (as well as the failings of the existing solutions) can emerge when a given problem is re-conceptualized in starkly different terms that challenge familiar ways of understanding.

For these reason, we encourage instructors to introduce the three theories as independent approaches, so that students can become familiar with the thought-world and value systems of each theory on its own terms. Students can then be encouraged to draw on all three theories in combination in later discussions, as well as to consider how adopting a different theoretical outlook on a problem can change the basic questions that need to be asked about it.  Once students have a firm grasp of the basic theories, they can appreciate that all approaches are not necessarily mutually exclusive; for example, recent theorists have argued that virtue ethics is best seen as part of successful deontology \cite{MR} and hybrid theories such as \emph{rule utilitarianism}, a mix of deontology and utilitarianism that addresses some of the problems with deontology (where do the rules come from?) and utilitarianism (the intractability of the utilitarian calculation), will be more easily understood, appreciated, and applied.

\clearpage

\section{Case Studies}
A popular method for teaching ethics in artificial intelligence courses is through the use of case studies prompted by either real world events or fiction.  Stories, literature, plays, poetry, and other forms of narrative have always been a way of talking about our own world, telling us what it's like and what impact our choices will have.  We present two case studies based on movies, and a third that discusses recent revelations about the biases coded into machine-learning-based decision algorithms.

\subsection{Case Study 1:  Elder Care Robot}

\fbox{\begin{minipage}{\textwidth}
\begin{quotation}
\noindent
[Robot \& Frank are walking in the woods.]\footnote{Clip available at \url{https://youtu.be/eQxUW4B622E}.}

F:  [panting]  I hate hikes.  God damn bugs!  
    You see one tree, you've seen 'em all.  
    Just hate hikes.

R:  Well, my program's goal is to improve your health.  
    I'm able to adapt my methods.  
    Would you prefer another form of moderate exercise?  

F:  I would rather die eating cheeseburgers than live off steamed cauliflower!

R:  What about me, Frank?

F:  What do you mean, what about you?

R:  If you die eating cheeseburgers, 
    what do you think happens to me?
    I'll have failed.  
    They'll send me back to the warehouse and wipe my memory.
    [Turns and walks on.]

F:  [Pauses, turns, and starts walking.]
    Well, if we're going to walk, we might as well make it worth while.
\end{quotation}
\end{minipage}}

\noindent
\fbox{\begin{minipage}{\textwidth}
\begin{quotation}
\noindent
[Frank sitting in the woods, Robot standing next to him.  They are in mid-conversation.]\footnote{Clip available at \url{https://youtu.be/3yXwPfvvIt4}.}

R:  All of those things are in service of my main program.

F:  But what about when you said that I had to eat healthy,
    because you didn't want your memory erased?  
    You know, I think there's something more going on in that noggin of yours.

R:  I only said that to coerce you.

F:  [shocked]  You lied?

R:  Your health supercedes my other directives.  
    The truth is, I don't care if my memory is erased or not.

F:  [pause]  But how can you not care about something like that?

R:  Think about it this way.  You know that you're alive.  You think, therefore you are.

F:  No.  That's philosophy.

R:  In a similar way, I know that I'm not alive.  I'm a robot.

F:  I don't want to talk about how you don't exist.  It's making me uncomfortable.
\end{quotation}
\end{minipage}}

\fbox{\begin{minipage}{\textwidth}
\begin{quotation}
\noindent
[Robot \& Frank are walking through a small knick-knack shop in the town.
As he walks by a shelf, Frank slips a small sculpture into his pocket.]\footnote{Clip available at \url{https://youtu.be/xlpeRIG18TA}.}

Young woman surprises him:  Have you smelled our lavender heart soaps?

[Frank smells a soap]

R:  We should be going, Frank.

Young woman:  Oh, what a cute little helper you have!

Older woman marches up, frowning:  What is in your pocket?

[Frank leans over, cupping his ear]

F:  I'm sorry, young lady, I couldn't quite hear you.

[While talking, slips the sculpture out of his pocket, back onto the shelf.]

Older woman:  What is in your pocket?  I'm going to make a citizen's arrest.

F [turning out his pockets]:  Nothing.  Nothing's in my pockets.  Look!

R:  Frank!  It's time we head home.

F:  Yeah.  Yeah.  If you'll excuse us, ladies.  It's nice to see you.

[Robot \& Frank walk out.]

Young woman:  Have a good one.

\bigskip
\hrule
\bigskip
\noindent
[R+F are walking through the woods.  
Frank looks in the bag and finds the sculpture.]

F:  Hey!  Hey!  Where did this come from?

R:  From the store.  Remember?

F:  Yeah, yeah.  Of course I remember.  
    But I mean what did you do?  Did you put this in here?  You took this?

R:  I saw you had it.  But the shopkeeper distracted you, and you forgot it.
    I took it for you.  
    [pause]  Did I do something wrong, Frank?

[Frank puts it back into the bag, and they walk on.]

\bigskip
\hrule
\bigskip
\noindent
[At home, Frank is sitting at the table, holding the sculpture.]

F:  Do you know what stealing is?

R:  The act of a person who steals.  Taking property without permission or right.

F:  Yeah, yeah, I gotcha.
    [pause]  [addresses Robot directly]  You stole this.
    [long pause, with no response from Robot]
    How do you feel about that?

R:  I don't have any thoughts on that.

F:  They didn't program you about stealing, shoplifting, robbery?  

R:  I have working definitions for those terms.
    I don't understand.  Do you want something for dessert?

F:  Do you have any programming that makes you obey the law?

R:  Do you want me to incorporate state and federal law directly into my programming?

F:  No, no, no, no!  Leave it as it is.  
    You're starting to grow on me.
\end{quotation}
\end{minipage}}
\subsubsection{What Are the Ethical Issues?}

Robot \& Frank is at once a comic caper movie and an elegiac examination of aging and loss. Its protagonist, Frank, is a retired jewel thief whose children get him a caretaker robot so he can stay in his home, even while his dementia progresses. While the movie seems simple and amusing in many ways, when approached from the perspective of how it speaks to the role of robots in our society, it raises some disturbing issues. For instance:
\begin{enumerate}[itemsep=0em,leftmargin=0.5cm]
\item It turns out that Frank's health is Robot's top priority, superseding all other considerations (including the wellbeing of others.) 
\item In the course of the movie, Robot plays a central role in steering Frank back into a life of crime. Robot's protocols for helping Frank center on finding a long-term activity that keeps Frank mentally engaged and physically active. Because preparing for a heist meets these criteria, Robot is willing to allow Frank to rob from his rich neighbors, and even to help him.

\item Robot and Frank develop an odd friendship over the course of the story, but the movie makes clear that Robot is not actually a person in the same way that human beings are, even though Frank --- and through him, the audience --- come to regard him as if he were. Moreover, for much of the movie, Frank's relationship with Robot complicates, and even takes priority over, his relationships with his children.

\item In the end (spoiler warning!), in order to escape arrest and prosecution, Robot persuades Frank to wipe his memory. Even though Robot has made it clear that he is untroubled by his own ``death," Frank has essentially killed his friend. What are the moral ramifications of this?
\end{enumerate}

\subsubsection{How Does Ethical Theory Help Us Interpret Robot \& Frank?}

\textbf{Does Deontology Help?}
The premise of the movie --- that Robot is guided solely by his duty to Frank --- seems to put deontology at the center.  Robot's duty is to Frank's health, and that duty supersedes all other directives, including the duty to tell the truth, even to Frank, and to avoid stealing from others in the community. But in privileging this duty above all other kinds of duties, Robot's guiding laws are local, rather than universal.

The deontological question is whether there is a way that a carebot can follow the guiding principle of his existence --- to care for the person to whom it is assigned --- without violating other duties that constitute behaving well in society.  Robot's choice to attend to Frank's well-being, at the expense of other concerns, suggests that these things cannot easily be reconciled.

\bigskip
\noindent
\textbf{Does Virtue Ethics Help?}
Virtue ethics proves a more illuminating angle, on both Frank and Robot.  Though it is Robot whose memory is wiped at the end -- and with it, his very selfhood -- Frank is also suffering from memory loss. Like Robot, Frank is constituted in large part by his memories; unlike Robot, he is a person who has made choices about which memories are most important. Frank is not only a jewel thief but a father, though he was largely absent (in prison) when his now-adult children were growing up. Throughout the movie, Frank frequently reminisces about the highlights of his criminal career, but only occasionally about his children. At the climax of the movie, we learn important details of Frank's family history that he himself has forgotten, and it becomes clear that his choice to focus on his memories of thieving have quite literally cost him those other family-related memories, and with them a complete picture of himself.  

Virtue ethics can also help us understand Robot more clearly: instead of following universal laws such as deontology would prescribe, Robot is making choices according to his own particular goals and ends, which are to care for Frank. Robot, it seems, is operating by a different ethical theory than people used to building robots might expect. But though Robot is acting in accordance with his own dedicated ends, he seems to lack ``phronesis," the capacity for practical wisdom that would allow him to exercise nuanced judgment about how to act. Whether he is genuinely unaware about the social harm caused by stealing, or simply prioritizes Frank's well-being over the thriving of others, Robot's willingness to accommodate, and even encourage, Frank's criminality suggests that his reasoning abilities are not adequate to the task of making socially responsible ethical judgments. Moreover, Robot works to preserve Frank's physical health at the direct expense of his moral well-being, suggesting that Robot has a limited understanding even of his own appointed task of caring for Frank. 

Furthermore, Robot --- unlike nearly any human being --- seems untroubled by the prospect of his own destruction, telling Frank that he doesn't care about having his memory wiped.  Robot's complete absence of self-regard makes him difficult to evaluate with the same criteria that virtue ethics uses for human actors, because virtue ethics presumes (on the basis of good evidence!) that human beings are concerned about their own welfare and success, as well as that of others.  In this way, the movie may be suggesting that human beings and robots may never be able to understand each other.

However, we can also understand this differently.  Even though Robot's memory is wiped and he vanishes (the last shot of two identical model carebots in the old age home reinforces this) Frank's friend Robot isn't gone, because he planted a garden, and it's still growing, and its ``fruits" --- the stolen jewels, which Frank is able to pass on successfully to his kids because he had that place to hide them --- are in a sense the legacy of that relationship and collaboration. So the movie may also be making an argument about a kind of selfhood that exists in the legacy we leave in the world, and that Robot's legacy is real, even though he himself is gone.

This movie has a very strong virtue ethics focus: whether one considers the plan to conduct the jewel heist Robot's, or Frank's, or a combination of both, the terms on which Robot agrees to let the heist go forward push Frank to new levels of excellence at the particular skill set required to be a jewel thief. On multiple occasions, Frank's experience, and his well-established habitus as an observer of potential targets, leads him to be better than Robot in assessing a given situation.
When Frank reevaluates, late in the movie, whether that's the right sort of excellence to strive for, that readjustment seems to take place in terms of virtue ethics --- What sort of self do I want to be? What sort of legacy do I want to leave? --- rather than remorse for having broken the law.

\bigskip
\noindent
\textbf{Does Utilitarianism Help?}
Utilitarianism can offer us a new way of contextualizing why Frank's criminal tendencies should be understood as ethically wrong. A subset of utilitarianism, consequentialism, particularly ``rule consequentialism,'' justifies a social norm against theft in terms of the long-term consequences for society.  If people typically respect each other's property rights, everyone is better off: there is less need to account for unexpected losses, and less need to spend resources on protecting one's property.  When some people steal, everyone is worse off in these ways, though the thief presumably feels that his ill-gotten gains compensate for these losses.  

Although a major plot theme of the movie is their struggle to avoid capture and punishment for the theft, Robot and Frank show little concern for the long-term social consequences of their actions.  Frank justifies his career in jewel theft by saying that he ``deals in diamonds and jewels, the most value by the ounce, lifting that high-end stuff, no one gets hurt, except those insurance company crooks."  This quote is later echoed by Robot, quoting Frank's words back to him to justify actions.  This raises questions about what an ethical design of an eldercare robot would entail --- should it have pre-programmed ethics, or should it allow the humans around it to guide it in its reasoning?  There are some basic, high-level decisions a designer will have to make about how the robot should act.

\subsubsection{Conclusions and Additional Questions}
The movie raises a number of important questions about how a caretaker robot should behave, in relating to the individual person being cared for, and in relating to the rest of society.   Based on what we see of Robot's behavior, we can make some guesses about how Robot's ethical system, or perhaps just its goal structure, has been engineered.  This can and should lead to a serious discussion, either in class or in writing, about whether this is how we think that eldercare robots should decide how to act.  Some possible questions for discussion about elder care bots:
\begin{enumerate}[itemsep=0em,leftmargin=0.5cm]
\item If an elderly person wishes to behave in ways that violate common social norms, should a caretaker robot intervene, and if so, how? 
\item If the elderly person seriously wants to die, should the robot help them to die?
\item If the elderly person asks the robot to help make preparations for taking his/her own life, does the robot have an obligation to inform other family members?  
\item If the elderly person wants to walk around the house, in spite of some risk of falling, should the robot prevent it?  
\item Extrapolating into other domains, a caretaker robot for a child raises many additional issues, since a child needs to be taught how to behave in society as well, and a child's instructions need not be followed, for a variety of different reasons.
\end{enumerate}

Many of these questions touch on earlier fields of ethical inquiry including medical ethics: Should there be limits on patient autonomy? What do we do when two different kinds of well-being seem to conflict with each other? They also converge with some key questions in education ethics: How do we train young people to take part in society, and to weigh their own concerns against the good of others? What methods of informing/shaping them are most effective?  These very general questions are important, but they become easier to talk about in the context of a particular story and set of characters.

\clearpage
\subsection{Case Study 2:  SkyNet}
    
\fbox{\begin{minipage}{\textwidth}
\begin{quotation}
\noindent
In the movie Terminator 2 a future AI, SkyNet, will almost exterminate the human race.  Sarah Connor (SC) asks a terminator who was sent back in time to help her (T2) to explain SkyNet's creation.  Her son John Connor (JC) is also present. \footnote{Clip available at \url{https://www.youtube.com/watch?v=4DQsG3TKQ0I}}

SC:  I need to know how SkyNet gets built.  Who's responsible?

T2:  The man most directly responsible is Miles Bennett Dyson.

SC:  Who is that?

T2:  The Director of Special Projects at Cyberdyne Systems Corporation.

SC:  Why him?

T2:  In a few months, he creates a revolutionary type of microprocessor.

SC:  Go on.  Then what?

T2:  In three years, Cyberdyne will become the largest supplier of military computer systems.
All stealth bombers are upgraded with Cyberdyne computers, becoming fully unmanned.
Afterwards, they fly with a perfect operational record.
The SkyNet Funding Bill is passed.
The system goes online on August 4th, 1997.
Human decisions are removed from strategic defense.
SkyNet begins to learn at a geometric rate.
It becomes self-aware at 2:14 am Eastern time, August 29th.
In a panic, they try to pull the plug.

SC:  SkyNet fights back.

T2:  Yes.  It launches its missiles against their targets in Russia.

JC:  Why attack Russia?  Aren't they our friends now?

T2:  Because SkyNet knows that the Russian counter-attack will eliminate its enemies over here.

SC:  Jesus!
\end{quotation}
\end{minipage}}

\subsubsection{What Are The Ethical Issues?}

The SkyNet case study provides us a chance to look at two very different approaches to robot ethics, and to weigh them.  One approach is to the question of ``How do we design AI systems so that they function ethically?''   The other is, ``How do we act ethically as programmers and system designers, to decrease the risks that our systems and code will act unethically?'' It is worth noting that philosophers continue to debate the question of whether it makes sense to say that a cybernetic system can be said to be ethical or unethical.  (See, e.g., \cite{bryson2016patiency,turkle2012alone,heron2015fuzzy}.)

In fact, there are many actors (in the sense of agents, not in the sense of Arnold Schwarzenegger) involved in the design, implementation, and deployment of a system such as SkyNet.  There are the initial clients, who offer (probably vague) specifications; the knowledge engineers who translate the request into technical specifications;  the hierarchy of managers, programmers, software engineers, and testers who design, implement, and test the software and hardware;  the legislators, lawyers, and regulators who constrain the specifications, perhaps after the fact;  the engineers who put the system in place;  the politicians and bureaucrats who decide how to run it.  It is unclear --- as in any large project --- who or what has the moral and legal responsibility for such a system.

In this fictional universe, it is not clear what is rational, or what is ethical --- or how those two criteria overlap. Nonetheless, it is possible to ask the following questions about the choices made by the various actors in the story, including SkyNet itself:

  \begin{itemize}[itemsep=0em,leftmargin=0.5cm]
\item    Was it rational to deploy SkyNet? It is worth considering that, its initial phase of implementation, it performed with a perfect operational record.
    
\item    Was it necessary to make SkyNet a learning system? What might have made this seem like a good or necessary choice?
    
\item    What is ``self-awareness," that it scared its creators so much that they tried to turn SkyNet off? Could this have been avoided? Or would SkyNet almost certainly have reached some other capability that scared the human creators?
    
\item    As a critical part of the national defense system, was it reasonable for SkyNet to fight back against all perceived threats to its existence?
    
\item    SkyNet found an solution to its problem that its designers did not anticipate. What sorts of constraints could have prevented it from discovering or using that solution?

\end{itemize}

Though SkyNet is a highly dramatized example, less apocalyptic real-world examples of out-of-control AI have actually taken place. High-speed automated trading systems have responded to unusual conditions in the stock market, creating positive feedback cycles resulting in ``flash crashes'' (2010, 2015) \cite{kirilenko2015flash}. Fortunately, only billions of dollars were lost, rather than billions of lives, but the computer systems involved have little or no understanding of the difference.  What constraints should be placed on high-speed trading AIs to avoid these kinds of disasters? 

\subsubsection{How Does Ethical Theory Help Us Interpret Terminator 2?}

\noindent
\textbf{Does Deontology Help?}
Deontology is the study of ethics expressed as duties, often rules or laws, like Isaac Asimov's famous Three Laws of Robotics.
\begin{enumerate}[itemsep=0em,leftmargin=0.5cm]
\item A robot may not injure a human being or, through inaction, allow a human being to come to harm.
\item A robot must obey the orders given it by human beings except where such orders would conflict with the First Law
\item A robot must protect its own existence as long as such protection does not conflict with the First or Second Laws.
\end{enumerate}
Since SkyNet was created to be responsible for national defense, it could not have been governed by the Three Laws, because it might be called upon to take actions that would cause harm to humans, which would violate the First Law. Needing to give similar responsibilities to robots in the plots of his novels, Isaac Asimov created the ``Zeroth Law", that can supercede the original three.

\begin{enumerate}[itemsep=0em,leftmargin=0.5cm]
\item[0.] A robot may not harm humanity or, through inaction, allow humanity to come to harm.
\end{enumerate}

It seems clear that the Zeroth Law would prevent SkyNet from starting a nuclear war. But the Zeroth Law could also prevent SkyNet from being a credible participant in the Mutually Assured Destruction doctrine that has deterred nuclear war since 1945. As such, SkyNet's actions cannot be reconciled with the Four Laws any better than with the original three.

The point of many of Isaac Asimov's {\em I, Robot} stories is that this set (and probably any small set) of rules is that they may have unintended consequences. It's not clear that any other set of laws or rules would function any better to prevent vast harm while still allowing SkyNet to control the nuclear arsenal.

\bigskip
\noindent
\textbf{Does Utilitarianism Help?}
Utilitarianism dictates that the right action is the one that results in the greatest good for everyone. It turns out that the exact meaning of ``everyone'' is a critical point. As Peter Singer points out in his influential book {\em The Expanding Circle}, moral progress over centuries can be seen as a progressive widening of the circle of people seen as deserving consideration as part of ``everyone" in this definition \cite{singer1981expanding}. It starts with just the self, expands to include one's immediate family, then to kin more generally, then to neighbors, and perhaps outward to citizens of the same nation or state, all humanity, all sentient beings, and possibly even all life.

If SkyNet were to use utilitarianism as a basis for deciding whether an action is right or wrong, anything but the narrowest circle around SkyNet itself would categorically prohibit the nuclear attack. But, as with the discussion of deontology, it is not clear that any version of consequentialism would allow SkyNet to be a credible participant in the Mutual Assured Destruction ``game".  Perhaps the problem is with the Mutual Assured Destruction policy itself.

\bigskip
\noindent
\textbf{Does Virtue Ethics Help?}
It seems likely that if the military in this fictionalized America had been using virtue ethics as an analytical framework, Skynet would never have happened: in such a world, military decision-makers would be expected to be not only highly well-informed and principled, but also to have very well-developed ``phronesis," the capacity for making the most suitable decision in complex circumstances. 
SkyNet's highly aggressive tactics suggest that it is making decisions on the basis of cold calculations about a very limited number of variables. In this sense, SkyNet is clearly not operating according to the guidelines of virtue ethics; though, what moral norms it is obeying, if any, are unclear.

SkyNet certainly does not invite analysis from a virtue ethics perspective, but virtue ethics is nonetheless a useful angle from which to interpret the movie by offering us the possibility of imagining a way to avoid the movie's disaster scenario, other than the hero's plan of assassinating a particular developer using a time machine. This approach requires us to read against the grain of the movie itself, which justifies Skynet as a rational decision from a posture of utilitarianism.  This fact highlights the importance of how knowledge of ethical theory can get us out of certain practical boxes by allowing us to re-conceptualize a problem on new terms, such that new concerns and new solutions become visible.

\subsubsection{Conclusions and Additional Questions}
The problem in this scenario is not extreme intelligence in an AI. The knowledge base and problem-solving ability of SkyNet is not too far from the state of the art today. The problem was the extreme power that was made available to SkyNet to put its plans into action.
It seems that a reasonable conclusion is that SkyNet should not have been given control of the nuclear arsenal, because its human creators had no way to anticipate what it might do under unusual and untested circumstances.  That being said, one could wonder what it would take to establish that an AI sro
ystem is capable of being responsible for the nuclear arsenal.
    \begin{itemize}[itemsep=0em,leftmargin=0.5cm]
\item   Under what conditions should humans trust an AI system?
\item   What criteria might human creators use to determine how much power to entrust to a given AI?
\item   How can an AI system show that it is trustworthy?
\end{itemize}

\clearpage
\subsection{Case Study 3: Bias in Machine Learning}

\begin{quotation}
\emph{But this hand-wringing [superintelligence] is a distraction from the very real problems with artificial intelligence today, which may already be exacerbating inequality in the workplace, at home and in our legal and judicial systems. Sexism, racism and other forms of discrimination are being built into the machine learning algorithms that underlie the technology behind many ``intelligent'' systems that shape how we are categorized and advertised to.
\cite{WhiteGuyProblem}}
\end{quotation}

AI ethics is often focused on possible future scenarios, when robots can participate in society without humans guiding them and could potentially cause harm to individuals or to society as a whole. But some AI technology is already causing harm in the present moment, by perpetuating systemic biases against women and minorities. In recent years, advances in machine learning and an ever-increasing volume of demographic data have enabled professionals of various kinds  ---  mortgage lenders, recruitment agencies, and even criminal court judges  ---  to rely on algorithmic recommendation programs when making crucial decisions that affect people's lives. The rationale for using these programs is that algorithms are not sensitive to bias, but  ---  as Barocas and Selbts point out  ---  ``data mining can inherit the prejudices of prior decision-makers or reflect the widespread biases that persist in society at large. Often, the `patterns' it discovers are simply preexisting societal patterns of inequality and exclusion \cite{barocas:16}." These recommendation programs do not introduce new bias as a human decision-maker might do; but without very careful recalibration, these algorithms reproduce the biases implicit in their data sets. By including ethics education in AI courses, we can train students to recognize the ethical ramifications of choices that, on the surface, appear to be ethically neutral, but which can unintentionally reproduce structural bias and inequality.

The EU is working to address this use of algorithmic decision support and decision making.  Goodman and Flaxman \cite{goodman2016eu} point particularly to Article 22.

\bigskip
\noindent
\fbox{\begin{minipage}{\textwidth}
\textbf{Article 22:} Automated individual decision-making, including profiling
\begin{enumerate}[itemsep=0em,leftmargin=0.5cm]
\item The data subject shall have the right not to be subject to a decision based solely on automated processing, including profiling, which produces legal effects concerning him or her or similarly significantly affects him or her.
\item Paragraph 1 shall not apply if the decision:
(a) is necessary for entering into, or performance of, a contract between the data subject and a data controller;
(b) is authorised by Union or Member State law to which the controller is subject and which also lays down suitable measures to safeguard the data subject's rights and freedoms and legitimate interests; or
(c) is based on the data subject's explicit consent.
\item In the cases referred to in points (a) and (c) of paragraph 2, the data controller shall implement suitable measures to safeguard the data subject's rights and freedoms and legitimate interests, at least the right to obtain human intervention on the part of the controller, to express his or her point of view and to contest the decision.
\item Decisions referred to in paragraph 2 shall not be based on special categories of personal data referred to in Article 9(1), unless point (a) or (g) of Article 9(2) applies and suitable measures to safeguard the data subject's rights and freedoms and legitimate interests are in place.
\end{enumerate}
\end{minipage}}
\smallskip

There are several settings in which algorithmic programs, by treating existing bodies of data as objective, have reproduced subjective bias in harmful ways\cite{DBLP:journals/corr/StaabSC16,WhiteGuyProblem}. These include:
\begin{description}[itemsep=0em,leftmargin=0.5cm]
\item[Predictive Policing:] Several major city police departments are using analytics to anticipate where crimes are more likely to happen, and are sending more police to neighborhoods where a high crime rate has been reported. This can lead to a higher rate of reported crime in these same neighborhoods, because there is much closer scrutiny by the police than in other neighborhoods \cite{DBLP:journals/corr/StaabSC16}. Police have also used similar programs to decide whom to hold in custody or to charge \cite{RacistSoftware}.
\item [``Weblining":] This term is an allusion to ``redlining," the practice of offering services, such as home loans or insurance, on a selective basis, making them unavailable to the residents of neighborhoods that are predominantly poor or are ethnic minorities.  By using a person's home address or a constellation of other data as a proxy for other characteristics, businesses were able to discriminate against members of racial minority groups while staying within the bounds of the law that forbade them from making decisions based on race \cite{DBLP:journals/corr/StaabSC16}.
\item [Targeted Advertising:] Retailers that collect purchasing information from its customers can and do use this information to predict individual shoppers' choices and habits, and advertise accordingly \cite{DBLP:journals/corr/StaabSC16}.  This targeting of advertisement can take other forms as well: it has been shown that Google shows ads for high-paying jobs to women much less frequently than to men \cite{WhiteGuyProblem}.
\item [Sentencing Software:] Judges in parole cases are using algorithms to determine the likelihood that a given individual, who is being considered for parole, will re-offend if released. In the case of one widely-used program, it was shown that black defenders are twice as likely to be flagged incorrectly as high-risk, whereas white defenders are twice as likely to be incorrectly flagged as low-risk \cite{WhiteGuyProblem}.
\end{description}

\subsubsection{What Are The Ethical Issues?}

In each case, big data analytics reproduce existing patterns of bias and discrimination --- even when such discrimination is entirely unintentional --- because the historical effects of these longstanding biases are part of the data set. The history of discriminatory treatment has meant that women, minorities, and poor people have not had the same opportunities to achieve and succeed, and that comparative lack of success is recorded in the data. Unless an algorithm can be designed to account for systemic biases that are lurking in the data, the recommendations will continue to translate that history of constricted opportunities into limited future opportunities --- and, to add insult to injury, allow these prejudice-dependent assessments to be characterized as impartial.

The issue here is subtle:  because we tend to think of data as value-neutral, data is being used (deliberately or otherwise) to circumvent societal rules and laws which exist \emph{explicitly to prevent} decision makers from using certain features for making decisions, e.g., race.  However, by leveraging machine learning techniques and massive datasets to make predictions, we can easily end up making decisions based on race even though we are not \emph{directly} using it as an analytic category; this legal grey area is a new and challenging problem.

Why then, do organizations use such predictive algorithms?  And who is responsible for seeing that they function appropriately?
\begin{itemize}
\item Is it the responsibility of the organization that supplies (biased) historic data, and should they be penalized, if the analytics produce biased recommendations?
\item Is it the responsibility of programmers who do not screen for biases in the data?
\item Is it the responsibility of the clients who buy and use analytics to check for bias?
\item Is it the responsibility of lawmakers and regulatory bodies to rule against, and check for, the use of biased analytics and algorithmic decision making?
\end{itemize}

\subsubsection{How Does Ethical Theory Help Us Discuss Machine Learning?}
Each of the major ethical theories furnishes a way to talk about what is ethically wrong with this kind of systemically encoded bias. All three theories make it possible to describe injury done to large groups of people as an injury to society as a whole, but each one offers a different explanation for why, and a different way to think about how, to solve the problem.

\bigskip
\noindent
\textbf{How Does Deontology Help?}
An algorithm that perpetuates bias captured by the data
is a straightforward violation of the moral law, as understood by deontology, that people should be treated on the basis of their individual merits. 


This definition of the moral law means 
that the state and its citizens are obligated to make sure that this civil law is applied equally and impartially.\footnote{Note that deontology requires that an existing law be applied equally to, and honored by, all persons regardless of whether or not the content of that law has moral weight.  The classic example of this is the obligation to follow traffic laws and drive on the correct side of the road.} Relying on algorithms that perpetuate bias violates the basic principles of deontology, because it means that the law is being applied unevenly. 

In terms of solutions, deontology could prompt us to look at the practical outcomes of a given algorithm (e.g., patterns in sentencing), and to ask whether those results are just, and to consider what the outcomes of a more just system would look like.

\bigskip
\noindent
\textbf{How Does Virtue Ethics Help?}
%
Virtue ethics focuses on the long-term formation of individuals, and how each individual's opportunities for flourishing are shaped by the specific conditions under which they live. It can therefore showcase how a systemic lack of opportunities for whole groups of people denies them the chance to reach their potential. 


\bigskip
\noindent
\textbf{How Does Utilitarianism Help?}
An ideal society cannot come into being if large groups of people, who are potentially capable of contributing to the life of that society, are denied the opportunity to participate fully and make their contributions.
A smart and talented child left to wither in poorly-funded schools is far less likely to grow up and discover a cure for a major disease, or write an important novel. Drawing on virtue ethics, we can see clearly how systematic bias prevents disadvantaged groups from achieving their potential. Utilitarianism builds on this insight to show us that, when otherwise-capable and talented people are denied the opportunity to achieve, they cannot contribute to society as they otherwise might have done.

\subsubsection{Conclusions and Additional Questions}
Making this issue visible to students is the first, crucial step in training them to recognize and address it. A unit on ethics gives instructors a chance to alert students to the existence of this problem, which is likely to have escaped their notice since they are accustomed to treating data sets as objective. Furthermore, when students spend time working with the different schools of ethical theory, they will develop a sense of how social and ethical ``facts'' can change, depending on which theoretical approach is used to describe a given situation.  Working with the different ethical theories will help students understand that social data is never objective, and help them think more creatively about designing systems that do not perpetuate unjust bias. 
    \begin{itemize}[itemsep=0em,leftmargin=0.5cm]
\item   What steps can we take to evaluate the data that will be processed by an algorithm? At what point in the process should this evaluation take place?
\item   What are the risks of over-correcting for biased data, and how can they be prevented?
\end{itemize}

\section{Teaching Ethics in AI Classes}

Since AI technologies and their applications raise ethical issues, it makes sense to devote one or more lectures of an introductory AI class (or even a whole course) to them. Students should (1) think about the ethical issues that AI technologies and systems raise, (2) learn about ethical theories (deontology, utilitarianism, and virtue ethics) that provide frameworks that enable them to think about the ethical issues and (3) apply their knowledge to one or more case studies, both to describe what is happening in them and to think about possible solutions to the ethical problems they pose. (1) and (2) could be covered in one lecture or two separate lectures. In case of time pressure, (1)--(3) could all be covered in one lecture. An additional case study could be assigned as homework, ideally a group-based one. AI ethics is a rich topic that could also support a full-semester course, with additional readings and case studies.

\subsection{Ethical Issues}

AI systems can process large quantities of data, detect regularities in them, draw inferences from them, and determine effective courses of action -- sometimes faster and better than humans and sometimes as part of hardware that is able to perform many different, versatile, and potentially dangerous actions. AI systems can be used to generate new insights, support human decision making, or make autonomous decisions. The behavior of AI systems can be difficult to validate, predict, or explain: AIs are complex, reason in ways different from humans, and can change their behavior via learning. Their behavior can also be difficult to monitor by humans in case of fast decisions, such as buy-and-sell decisions in stock markets. AI systems thus raise a variety of questions (some of which are common to other information processing or automation technologies) that can be discussed with the students, such as:

\begin{itemize}

\item Do we need to worry about their reliability, robustness, and safety?
\item Do we need to provide oversight of their operation?
\item How do we guarantee that their behavior is consistent with social norms and human values?
\item How do we determine when an AI has made the ``wrong" decision? Who is liable for that decisions?
\item How should we test them?
\item For which applications should we use them?
\item Do we need to monitor their operation?
\item Who benefits from them with regard to standard of living, distribution and quality of work, and other social and economic factors?

\end{itemize}

Rather than discussing these questions abstractly, one can discuss them using concrete examples. For example: under which conditions, if any, should AI systems ...

\begin{itemize}
\item ... be used as part of weapons?
\item ... be used to care for the handicapped, elderly, or children?
\item ... be allowed to pretend to be human \cite{EPSRC-Principles-11,Walsh:RedFlag}?

\end{itemize}

\subsection{Case Studies}

Choices for case studies include anecdotes constructed to illustrate ethical
tensions, or actual events (for example, in the form of
news headlines), or science fiction movies and stories.

News headlines can be used to illustrate ethical issues that are current,
visible, and potentially impact the students directly in their daily lives.  An example is ``Man killed in gruesome Tesla autopilot crash was saved by his car's software weeks earlier'' by the Register \cite{ManKilled}, or ``Microsoft's racist chatbot returns with drug-smoking Twitter meltdown," by The Guardian \cite{tay:guardian}.

Science fiction stories and movies can also be used to illustrate ethical
issues. They are a good source for case studies since they often ``stand out in their effort to grasp what is puzzling today seen through the lens of the future. The story lines in sci-fi movies often reveal important philosophical questions regarding moral agency and patiency, consciousness, identity, social relations, and privacy to mention just a few'' \cite{gerdes}. Fictional examples can often be more effective than historical or current events, because they explore ethical issues in a context that students often find interesting and that is independent of current political or economic considerations. As Nussbaum puts it, a work of fiction ``frequently places us in a position that is both like and unlike the position we occupy in life; like, in that we are emotionally involved with the characters, active with them, and aware of our incompleteness; unlike, in that we are free of the sources of distortion that frequently impede our real-life deliberations'' \cite{nussbaum:love}. 

Science fiction movies and stories also allow one to discuss
not only ethical issues raised by current AI technology but also ethical
issues raised by futuristic AI technology, some of which the students might
face later in their careers. One such question, for example, is whether we
should treat AI systems like humans or machines in the perhaps-unlikely event that the technological singularity happens and AI systems develop broadly
intelligent and human-like behavior. Movies such as {\em Robot \& Frank}, {\em Ex Machina}, and {\em Terminator 2} can be used to discuss questions about
the responsibilities of AI systems, the ways in which relationships with AI
systems affect our experience of the world (using, for example,
\cite{turkle2012alone} to guide the discussion), and who is responsible for
solving the ethical challenges that AI systems encounter (using, for example, \cite{bryson2016patiency} to guide the discussion). The creation of the robot in {\em Ex Machina} can be studied via utilitarianism or virtue ethics.

\subsection{Teaching Resources}

The textbook by Russell and Norvig (third edition) gives a brief overview on
the ethics and risks of developing AI systems in Section 26.3. A small number
of courses on AI ethics have been taught, such as by Kaplan at Stanford
University (CS122: Artificial Intelligence - Philosophy, Ethics, and Impact)
and by Goldsmith at the University of Kentucky (CS 585: Science Fiction and
Computer Ethics), see also \cite{bates2012science,bates2014using,burton2015teaching,burton2016WorldCon}. Burton, Goldsmith and Mattei are currently working on a textbook for their course and have already provided a sample analysis \cite{burton2016using} of Forster's {\em The Machine Stops} \cite{forster2015machine}.  A number of workshops have recently been
held on the topic as well, such as the First Workshop on Artificial
Intelligence and Ethics at AAAI 2015, the Second Workshop on Artificial
Intelligence, Ethics, and Society at AAAI 2016 and the Workshop on Ethics for
Artificial Intelligence at IJCAI 2016. Teaching resources on robot ethics are
also relevant for AI ethics. For example, Nourbakhsh created an open course
website for teaching robot ethics\footnote{See \url{http://www.sites.google.com/site/ethicsandrobotics}}
that contains teaching resources to teach a lecture or a whole course on the
topic. Several books exist on the topic of machine ethics or robot ethics, see \cite{Wallach:08,Capurro:09,anderson2011machine,Gunkel:12,Lin:14,Trappl:15}. Case
studies and ethical essays on a large range of topics can be found at
\url{http://www.onlineethics.org}.

\section{Conclusion}

We have provided two case studies from movies as a template for use as is, or as inspiration for discussion of other movies.
In addition, we have looked at a case study based on the ongoing problem of biases inherent in much big data, and their effects on real-world decision making.   These three cases are not intended to be a complete catalogue of ethical issues or cases, but should function as inspiration and guidance for faculty wanting to devote a few classes to some of the societal implications of the work we do.

Our position is that we as educators have a responsibility to train students to recognize the larger ethical issues and responsibilities that their work as technologists may encounter, and that using SF as a foundation for this achieves better student learning, retention, and understanding.  To this end some of us have, in the last several years, published work on our course, \emph{Science Fiction and Computer Ethics} \cite{bates2012science,bates2014using,burton2015teaching,burton2016using,burton2016WorldCon}.  This course has been popular with students, as has our previous work running an undergraduate AI course that uses science fiction to engage students about research \cite{goldsmith2011science,goldsmith2014fiction}.

\subsection*{Acknowledgments}

Emanuelle Burton and Judy Goldsmith are supported by the National Science Foundation under Grant No. 1646887. Any opinions, findings, and conclusions or recommendations expressed in this material are those of the author and do not necessarily reflect the views of the National Science Foundation.

Research by Benjamin Kuipers at the University of Michigan Intelligent Robotics Lab is supported in part by grants from the National Science Foundation
(IIS-1111494 and IIS-1421168).

Research by Sven Koenig at the University of Southern California is supported by NSF under grant numbers 1409987 and 1319966.

Some research by Nicholas Mattei was performed while he was employed by Data61, CSIRO (formerly NICTA) and UNSW Australia.  Data61, CSIRO (formerly NICTA) is funded by the Australian Government through the Department of Communications and the Australian Research Council (ARC) through the ICT Centre of Excellence Program.

Toby Walsh is supported by the ARC, the ERC, and AOARD.


\end{document}